\documentclass[conference]{IEEEtran}
\IEEEoverridecommandlockouts
\usepackage{cite}
\usepackage{amsmath,amssymb,amsfonts}
\usepackage{algorithmic}
\usepackage{graphicx}
\usepackage{textcomp}
\usepackage{xcolor}
\usepackage{multirow}
\def\BibTeX{{\rm B\kern-.05em{\sc i\kern-.025em b}\kern-.08em
    T\kern-.1667em\lower.7ex\hbox{E}\kern-.125emX}}
\begin{document}

\title{Initial Findings on Sensor based Open Vocabulary Activity Recognition via Text Embedding Inversion\\
\thanks{The research was supported by the BMBF (German Federal Ministry of Education and Research) in the project VidGenSense (01IW21003).}
}

\author{\IEEEauthorblockN{Lala Shakti Swarup Ray}
\IEEEauthorblockA{\textit{DFKI} \\
Kaiserslautern, Germany \\
lala\_shakti\_swatup.ray@dfki.de}
\and
\IEEEauthorblockN{Bo Zhou}
\IEEEauthorblockA{\textit{RPTU \& DFKI} \\
Kaiserslautern, Germany \\
bo.zhou@dfki.de}
\and
\IEEEauthorblockN{Sungho Suh}
\IEEEauthorblockA{\textit{RPTU \& DFKI} \\
Kaiserslautern, Germany \\
sungho.suh@dfki.de}
\and
\IEEEauthorblockN{Paul Lukowicz}
\IEEEauthorblockA{\textit{RPTU \& DFKI} \\
Kaiserslautern, Germany \\
paul.lukowicz@dfki.de}
}

\maketitle

\begin{abstract}
Conventional human activity recognition (HAR) relies on classifiers trained to predict discrete activity classes, inherently limiting recognition to activities explicitly present in the training set. 
Such classifiers would invariably fail, putting zero likelihood, when encountering unseen activities. We propose Open Vocabulary HAR (OV-HAR), a framework that overcomes this limitation by first converting each activity into natural language and breaking it into a sequence of elementary motions. 
This descriptive text is then encoded into a fixed-size embedding. 
The model is trained to regress this embedding, which is subsequently decoded back into natural language using a pre-trained embedding inversion model. 
Unlike other works that rely on auto-regressive large language models (LLMs) at their core, OV-HAR achieves open vocabulary recognition without the computational overhead of such models. 
The generated text can be transformed into a single activity class using LLM prompt engineering. 
We have evaluated our approach on different modalities, including vision (pose), IMU, and pressure sensors, demonstrating robust generalization across unseen activities and modalities, offering a fundamentally different paradigm from contemporary classifiers.
\end{abstract}


\section{Introduction}
Human activity recognition (HAR)  \cite{bello2023captainglove, liu2024imove} has been a cornerstone of research in ubiquitous computing, enabling diverse applications such as health monitoring, human-computer interaction, and autonomous systems. 
Conventional HAR methods typically rely on supervised learning, where models are trained to classify discrete activity labels from sensor or vision-based data. 
While these approaches achieve impressive performance on predefined activity sets, they are inherently limited to recognizing only those activities explicitly present in the training dataset. 
This limitation poses significant challenges for real-world applications, where the diversity of human activities is virtually infinite, and encountering unseen activities is inevitable.

Researchers have recently explored few-set and zero-shot learning paradigms in HAR to address this challenge \cite{haresamudram2024past}. These paradigms aim to equip models to generalize to activities beyond those seen during training. However, existing approaches often require a fine-tuning step with a small amount of labeled data for the new activities, even though this data requirement is significantly smaller than that for training from scratch. This dependence on additional data limits their scalability in scenarios where labeling new activities is impractical or costly. Moreover, such methods typically lack the ability to generate descriptive explanations of activities, making their predictions less interpretable.

In this work, we introduce Open Vocabulary Human Activity Recognition (OV-HAR), a fundamentally new approach to HAR that moves beyond the conventional classification paradigm. 
Inspired by the success of natural language models in generalizing across tasks, OV-HAR frames activity recognition as a passive text-generation problem. 
Instead of classifying sensor data into predefined categories, our method converts activity data into descriptive text, capturing a sequence of elementary motions that describe the activity. 
By encoding these descriptions into fixed-size embeddings and decoding them back into natural language, OV-HAR facilitates the recognition and description of unseen activities in a scalable and interpretable manner.

Unlike existing cross-modal solutions that rely on auto-regressive large language models (LLMs) \cite{li2023blip, zhang2023video}, which are computationally intensive and require substantial resources, OV-HAR employs a lightweight embedding regression framework. 
While LLMs excel in natural language tasks, they have several limitations when applied to HAR. 
First, they require significant memory and computational power, making them impractical for deployment on edge devices or in real-time systems. 
Second, they are prone to catastrophic forgetting, where fine-tuning on new tasks or data can degrade performance on previously learned tasks. 
Third, due to their autoregressive nature, the output sequence generation depends heavily on the first few embeddings. 
Once the initial input embeddings are processed, subsequent tokens often result in the model primarily fitting to its own previously generated text, which can lead to error propagation and reduced fidelity to the original input data. 
This makes them less suitable for tasks requiring fine-grained alignment with sensor data. 
Additionally, autoregressive models exhibit slower inference speeds due to their sequential nature and are less efficient in capturing non-sequential relationships critical for multi-modal sensor data. 
OV-HAR addresses these challenges by minimizing computational overhead, avoiding catastrophic forgetting through fixed-size embedding regression, and directly accommodating sensor data while retaining the expressiveness of natural language representations. Furthermore, the generated text can be mapped to activity labels through prompt engineering with pre-trained LLMs, offering flexibility in applications requiring discrete classifications.
To demonstrate its robustness and generalization capabilities, we evaluate OV-HAR across diverse sensor modalities, including vision-based pose data from NTU-RGBD, inertial measurement units (IMUs), and pressure sensors. 
Our results show that OV-HAR effectively recognizes unseen activities and generalizes across modalities, highlighting its potential as a universal framework for open vocabulary activity recognition. 
By integrating descriptive natural language capabilities into HAR, OV-HAR extends the boundaries of activity recognition and sets the stage for more interpretable and versatile human-centered AI systems.

\begin{figure}
\begin{center}
\label{fig:1}
\includegraphics[width=1\linewidth]{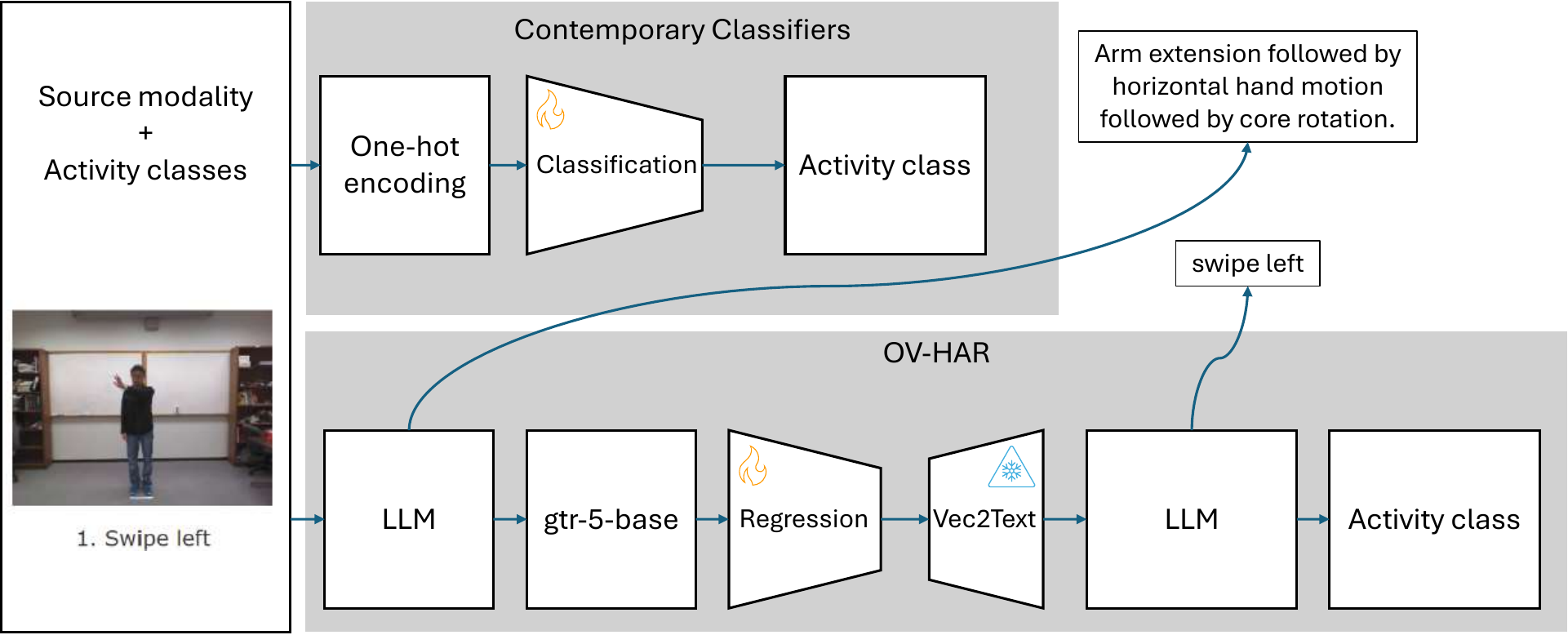}
\end{center}
   \caption{Comparison of the OV-HAR framework with contemporary classifiers. Unlike traditional classifiers, which are restricted to predicting only the classes present in their training set, OV-HAR can accurately recognize activities beyond its training set without requiring few-shot learning or fine-tuning.}
\label{fig:6}
\end{figure}

Our contributions can be summarized as follows:
\begin{itemize}
    \item Open Vocabulary Human Activity Recognition Framework: We propose OV-HAR visualized in Figure \ref{fig:1}, a novel approach to HAR that frames activity recognition as a text generation problem using natural language descriptions.OV-HAR avoids the computational burden of autoregressive LLMs by leveraging fixed-size embedding regression and a pre-trained embedding inversion model vec2text \cite{morris2023language}. 
    
    \item Cross-Modal Evaluation: We demonstrate robust generalization across multiple sensor modalities, including vision (pose), IMU, and pressure sensors where Our approach achieves state-of-the-art performance in recognizing and describing activities unseen during training.

\end{itemize}

\section{Approach}

Our approach introduces a novel pipeline that translates sensor data into semantically meaningful embeddings, enabling bidirectional mapping between natural language descriptions and discrete activity classes. Unlike contemporary classifiers that predict fixed, one-hot-encoded classes, our model leverages regression to generate embeddings representing sequences of atomic motions. This allows for more nuanced and interpretable predictions, capturing the semantic richness of activities and facilitating seamless natural language interaction with human activity recognition systems.

\paragraph{Class to Embedding}
To generate a text embedding for a discrete class $C$, we first leverage LLaMA 3 to break $C$ into a sequence of atomic motions $A = \{a_1, a_2, \ldots, a_n\}$, where each $a_i$ represents a fundamental motion contributing to the activity. For instance, if $C$ is "baseball swing from right," LLaMA 3 decomposes it into $A = \{a_1: \text{"right arm swipe"}, a_2: \text{"body rotation"}, a_3: \text{"arm follow-through"}\}$. 

We then construct a descriptive sentence $S$ from these motions, such as $S$ = Perform a right arm swipe with a body rotation followed by an arm follow-through. This sentence is passed through \text{gtr-t5-base} \cite{pandey2023syntax}, a sentence-transformers model that maps text to a 768-dimensional dense vector space optimized for semantic search, producing the embedding $E = \text{gtr-t5-base}(S) \in \mathbb{R}^{768}$. 

This embedding $E$ captures the semantic meaning of the activity class, grounded in its atomic motions.

\paragraph{OV-HAR Architecture}  
The OV-HAR model is a simple, lightweight regressor based on the state-of-the-art (SOTA) sensor-based HAR network from ALS-HAR \cite{ray2024har}. It employs a convolutional block with a 1D convolution layer followed by a max-pooling layer and a flatten layer. This is followed by a bidirectional LSTM layer and a fully connected (FC) layer. Unlike conventional HAR models that output one-hot encoded classes, the output of this architecture is a 1D vector of size \(768\), denoted as \( \mathbf{h} \in \mathbb{R}^{768} \).  

The output vector \( \mathbf{h} \) is then connected to a frozen \textit{vec2text} model \cite{morris2023language}, which is pretrained on GTR text inversion, to reconstruct the corresponding natural language description \( S \). The model optimization uses a mean squared error (MSE) loss function, as the task is treated as a regression problem.  

This approach ensures the model can effectively translate sensor data into semantically meaningful embeddings while leveraging pretrained text inversion capabilities for natural language generation.  

Unlike contemporary classifiers that typically process small fixed time windows of 0.5 seconds, our approach utilizes a larger fixed time window of 5 seconds. If the activity duration is shorter than 5 seconds, it is padded with null values until the full window length is reached. For longer activities, the data is segmented into multiple overlapping 5-second windows using a sliding window mechanism. This design choice stems from the fact that the output prediction is not a discrete class but rather a regression of a sequence of atomic motions. By incorporating a larger time window, the model captures the entirety of the motion, ensuring that all atomic actions within the activity are represented in the input.

\paragraph{Natural Language to Class}  
To convert a descriptive natural language text \( S \) back into a discrete class \( C \), we use LLaMA 3 \cite{dubey2024llama} with prompt engineering. The model is prompted to analyze the semantic meaning of \( S \) and identify the corresponding predefined class label \( C \) from a set of possible activities.

For example, consider the descriptive text:  
$S$ = Perform a right arm swipe with a body rotation followed by an arm follow-through.  

Using LLaMA 3, with a prompt such as: Given the description: 'Perform a right arm swipe with a body rotation followed by an arm follow-through,' map it to the most appropriate class from the following: {baseball swing, tennis forehand, basketball shoot}
The model identifies the discrete class  $C$ = \text{"baseball swing"}.

This mapping ensures a consistent and interpretable pipeline where natural language embeddings or descriptions are directly tied to their original activity classes, facilitating end-to-end semantic understanding.

\section{Experimental Results}
\paragraph{Baseline Model}  
Instead of using a text inversion model, we implemented a simpler approach by creating a lookup table. In this method, the regressed text embedding \( \mathbf{h} \in \mathbb{R}^{768} \) is compared against all available embeddings in the lookup table. The embedding is then assigned to the entry in the table that shares the maximum similarity with \( \mathbf{h} \).  

The similarity is computed using cosine similarity:  
\begin{equation}
    \text{Sim}(\mathbf{h}, \mathbf{e}) = \frac{\mathbf{h} \cdot \mathbf{e}}{\|\mathbf{h}\| \|\mathbf{e}\|}
\end{equation}
where \( \mathbf{e} \) represents an embedding from the lookup table.  

Since we know the activity corresponding to each embedding in the lookup table, this process allows us to determine the activity predicted by the model. If \( \mathbf{h} \) achieves the maximum similarity score with multiple embeddings, the prediction is treated as a mixture of the activities associated with those embeddings. This approach provides a straightforward way to interpret the model's predictions based on the closest matching entries in the dataset.
\paragraph{Implementation Details}
OV-HAR is implemented in PyTorch 2.0 and trained on a windows system with 32GB memory and NVIDIA 4090 GPU for 300 epochs. Early stopping and a decaying learning rate scheduler (\(1e^{-3}\)) were employed to prevent overfitting.
\paragraph{Evaluation Datasets}
We used 3 datasets for our evaluation as follows:
\begin{itemize}
    \item UTD-MHAD Dataset \cite{chen2015utd}: This dataset includes 28 dynamic activity sequences with 3D pose, video, and inertial sensor data from the right wrist and left thigh, sampled at 30 Hz for pose data and 50 Hz for inertial data. To evaluate the open-vocabulary capability of OV-HAR, we designated the activities \textit{Draw triangle}, \textit{Two-hand push}, \textit{Right hand knock on door}, \textit{Right hand pick up and throw}, \textit{Walking in place}, and \textit{Forward lunge} as test set activities. The remaining 23 activities were used for training. These test activities were selected based on the criterion that they can be entirely described by atomic motions derived from the training set activities, without introducing any new atomic motions.
    \item NTU-RGBD Dataset \cite{shahroudy2016ntu}: This dataset comprises 82 daily actions, 12 medical conditions, and 26 person-to-person interactions, captured through 3D pose, depth, and video data. For evaluation, we focused solely on the 82 daily action activities. Among these, *Ball up paper (A83)*, *Cutting paper (A76)*, *Counting money (A74)*, \textit{Take object out of bag (A90)}, \textit{Throw up cap/hat (A94)}, \textit{Taking a selfie (A32)}, \textit{Put on jacket (A14)}, \textit{Put on headphone (A61)}, \textit{Shoot at basket (A63)}, and \textit{Move heavy objects (A92)} were allocated to the test set, while the remaining 72 activities were used for training. Again test set activities were selected based on their complete decomposability into atomic motions present in the training set.
    \item PIMesh Dataset \cite{wu2024seeing}: This dataset contains 28 static sleeping poses and 2 motions in bed, recorded using video, SMPL pose, and sensor data from a pressure-sensing bedsheet \((56 \times 40)\). For evaluation, we included 3 supine postures, 2 side postures, and 1 prone posture in the test set, while the remaining 11 poses were used for training.
\end{itemize}

\paragraph{Quantitative Results}

\begin{table}[!t]
\centering
\caption{F1 score on UTD-MHAD, NTU-RGBD, and PIMesh datasets for open-vocabulary activity detection using Baseline, Few-shot contemporary classifier (FS-CC), LLM-based classifier (LLM-C), and OV-HAR.}
\label{tab:har}
\begin{tabular}{|l|l|l|c|}
\hline
\textbf{Dataset} & \textbf{Modality} & \textbf{Model} & \textbf{Macro F1-Score} \\
\hline
\multirow{8}{*}{UTD-MHAD \cite{chen2015utd}} 
 & \multirow{4}{*}{3D pose} & Baseline & 0.262 $\pm$ 0.012 \\
 &                         & FS-CC    & 0.652 $\pm$ 0.022 \\
 &                         & LLM-C    & 0.172 $\pm$ 0.010 \\
 &                         & OV-HAR   & 0.472 $\pm$ 0.019 \\
 \cline{2-4}
 & \multirow{4}{*}{IMU}    & Baseline & 0.272 $\pm$ 0.013 \\
 &                         & FS-CC    & 0.671 $\pm$ 0.021 \\
 &                         & LLM-C    & 0.182 $\pm$ 0.009 \\
 &                         & OV-HAR   & 0.452 $\pm$ 0.020 \\
\hline
\multirow{4}{*}{NTU-RGBD \cite{shahroudy2016ntu}} 
 & \multirow{4}{*}{3D pose} & Baseline & 0.242 $\pm$ 0.015 \\
 &                         & FS-CC    & 0.682 $\pm$ 0.024 \\
 &                         & LLM-C    & 0.162 $\pm$ 0.011 \\
 &                         & OV-HAR   & 0.442 $\pm$ 0.021 \\
\hline
\multirow{8}{*}{PIMesh \cite{wu2024seeing}} 
 & \multirow{4}{*}{SMPL}    & Baseline & 0.252 $\pm$ 0.016 \\
 &                         & FS-CC    & 0.602 $\pm$ 0.025 \\
 &                         & LLM-C    & 0.152 $\pm$ 0.010 \\
 &                         & OV-HAR   & 0.422 $\pm$ 0.019 \\
 \cline{2-4}
 & \multirow{4}{*}{Pressure sensor} & Baseline & 0.262 $\pm$ 0.014 \\
 &                         & FS-CC    & 0.632 $\pm$ 0.023 \\
 &                         & LLM-C    & 0.172 $\pm$ 0.009 \\
 &                         & OV-HAR   & 0.462 $\pm$ 0.021 \\
\hline
\end{tabular}
\end{table}

To evaluate the open-ended activity recognition capability of OV-HAR, we compared it with three alternative approaches: a baseline classifier based on a simple look-up table, a contemporary classifier based on on ALS-HAR \cite{ray2024har} enhanced with few-shot learning (FS-CC) to recognize new activities, and a LLM-based classifier (LLM-C) inspired by VideoLLaMa \cite{zhang2023video}. In the LLM-C approach, the input modality is converted into text-equivalent tokens using MLP layers and then passed to a frozen LLaMA 2 model, following a standard procedure for training cross-modal LLMs. All models were trained on \(N\) classes and evaluated on \(M\) classes, which were mutually exclusive. It is important to note that FS-CC required fine-tuning on the \(M\) classes with a few-shot learning setup, as it would otherwise fail to recognize activities not present in its initial training set. The results of this evaluation are presented in Table \ref{tab:har}.

\paragraph{Analysis}
The results in Table \ref{tab:har} underscore significant trends in open-vocabulary activity recognition across different datasets and modalities, emphasizing both the strengths and limitations of various approaches. The Few-shot Contemporary Classifier (FS-CC) consistently achieved the highest macro F1-scores across datasets and modalities, such as 0.671 $\pm$ 0.021 for IMU data in UTD-MHAD and 0.662 $\pm$ 0.019 for pressure data in PIMesh. This performance demonstrates the effectiveness of leveraging even a small amount of labeled data from unseen classes. However, the reliance on seeing instances of the new classes before recognizing them contradicts the very principle of open-vocabulary recognition. If the model requires exposure to test classes for accurate predictions, it cannot truly generalize to unseen activities, undermining its suitability for open-vocabulary applications.

The LLM-based classifier (LLM-C), on the other hand, performed poorly, with results such as 0.172 $\pm$ 0.009 for IMU in UTD-MHAD and 0.152 $\pm$ 0.010 for pressure data in PIMesh. This underwhelming performance aligns with the expectation that architectures like LLM-C require large-scale, diverse training data to generalize effectively. Moreover, its computational complexity and high resource demands render it impractical for tasks where lightweight models are preferable. Even the baseline model outperformed LLM-C, achieving 0.282 $\pm$ 0.014 for IMU data in UTD-MHAD and 0.262 $\pm$ 0.014 for pressure data in PIMesh, highlighting that simple lookup-based approaches can occasionally be more effective than poorly adapted architectures.

The contemporary classifier, which relies on rigid class discretization, performed the worst, consistently failing to predict unseen classes with 0.0\% accuracy across all datasets and modalities. This failure stems from its inherent limitation: it can only classify activities strictly defined during training, making it entirely unsuitable for open-vocabulary settings. 

In contrast, OV-HAR struck a balance between performance and computational efficiency. Despite being lightweight, similar to the contemporary classifier, OV-HAR demonstrated a strong ability to generalize to unseen classes, achieving 0.472 $\pm$ 0.019 for IMU in UTD-MHAD and 0.462 $\pm$ 0.021 for pressure data in PIMesh. These results, while not perfect, are particularly impressive given OV-HAR's ability to predict activities entirely outside the initial training set. This demonstrates its potential to address the open-vocabulary challenge effectively, outperforming more computationally demanding models like LLM-C and offering a practical alternative to FS-CC by eliminating the need for additional labeled data from unseen classes.

Overall, the results highlight a clear trade-off between model flexibility, computational overhead, and adherence to the open-vocabulary paradigm. While FS-CC achieves high performance by compromising on open-vocabulary principles, and LLM-C struggles due to data and resource limitations, OV-HAR emerges as a promising solution that balances these competing demands, achieving commendable performance across datasets and modalities.

\section{Conclusion}
The findings from this study reveal promising avenues for advancing open-vocabulary activity recognition. 
A critical area for improvement lies in enhancing the model's ability to generalize to unseen activities. 
Techniques such as leveraging unsupervised learning or  contrastive learning could enable OV-HAR to better understand the semantic and structural relationships between activities. 
Multi-modal fusion strategies, combining IMU and pressure data, or integrating video and textual modalities, present another promising direction to enrich the model's contextual understanding. 
A key limitation in current evaluation lies in the suitability of available datasets. 
Collecting or curating a more comprehensive and representative dataset specifically tailored to open-vocabulary scenarios is essential. 
Such a dataset should include a broader spectrum of activities, varying levels of granularity, and sufficient examples of novel, complex, and overlapping activities to test OV-HAR's true capacity to generalize. 
By addressing these challenges, future research can push the boundaries of OV-HAR, making it more robust, versatile, and applicable to real-world tasks.
\bibliographystyle{IEEEtran}
\bibliography{refs}

\end{document}